\documentclass[10pt,twocolumn,letterpaper]{article}

\usepackage{cvpr}
\usepackage{times}
\usepackage{epsfig}
\usepackage{graphicx}
\usepackage{amsmath}
\usepackage{amssymb}
\usepackage{subcaption}
\usepackage[linesnumbered]{algorithm2e}
\usepackage{float}
\usepackage{caption, cuted}

\usepackage[breaklinks=true,bookmarks=false]{hyperref}

\cvprfinalcopy % *** Uncomment this line for the final submission

\def\cvprPaperID{****} % *** Enter the CVPR Paper ID here

\begin{document}

%%%%%%%%% TITLE
\title{Semi-parametric Image Inpainting}

\author{Karim Iskakov\\
MIPT\\
{\tt\small iskakov@phystech.edu}
}

% \and

\twocolumn[{%
\renewcommand\twocolumn[1][]{#1}%
\maketitle

\begin{center}
    \vspace*{-5mm}
    \centering
    \includegraphics[width=1.0\textwidth]{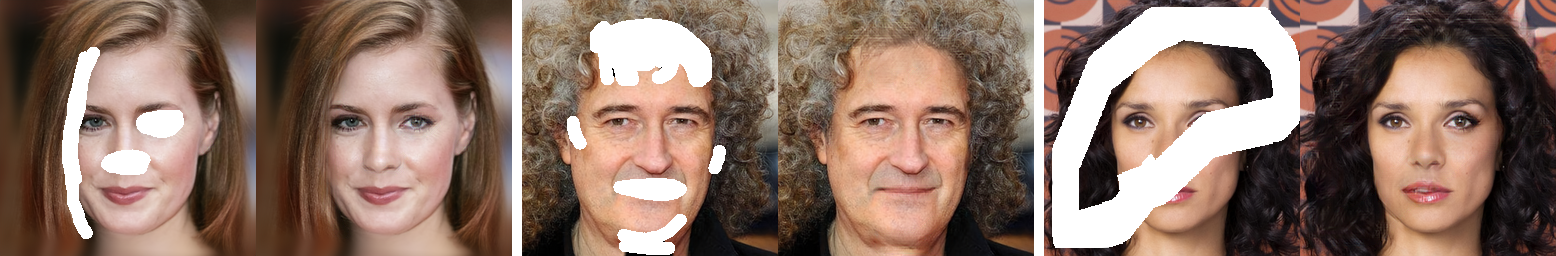}
    \vspace*{2mm}
\end{center}%
}]
%\thispagestyle{empty}

%%%%%%%%% ABSTRACT
\begin{abstract}
This paper introduces a semi-parametric approach to image inpainting for irregular holes. The nonparametric part consists of an external image database. During test time database is used to retrieve a supplementary image, similar to the input masked picture, and utilize it as auxiliary information for the deep neural network. Further, we propose a novel method of generating masks with irregular holes and present public dataset with such masks. Experiments on CelebA-HQ dataset show that our semi-parametric method yields more realistic results than previous approaches, which is confirmed by the user study. 
\end{abstract}

\begin{figure*}[h]
\captionsetup[subfigure]{justification=centering}
\begin{center}
  \begin{subfigure}[t]{0.13\textwidth}
      \includegraphics[width=\textwidth]{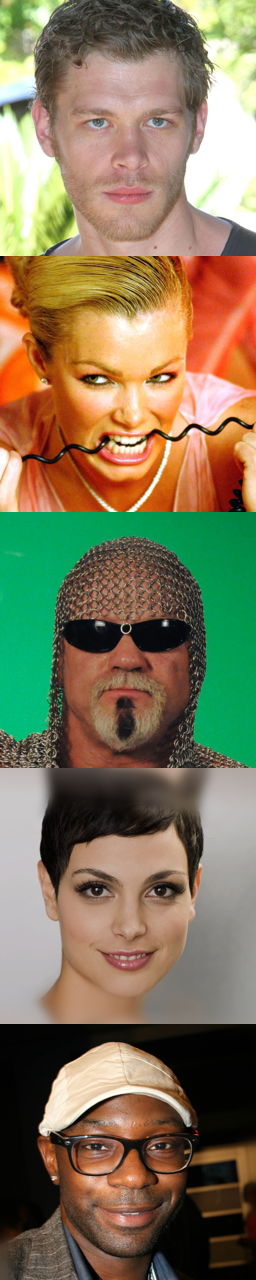}
    \caption{Original}
  \end{subfigure}%
  \begin{subfigure}[t]{0.13\textwidth}
      \includegraphics[width=\textwidth]{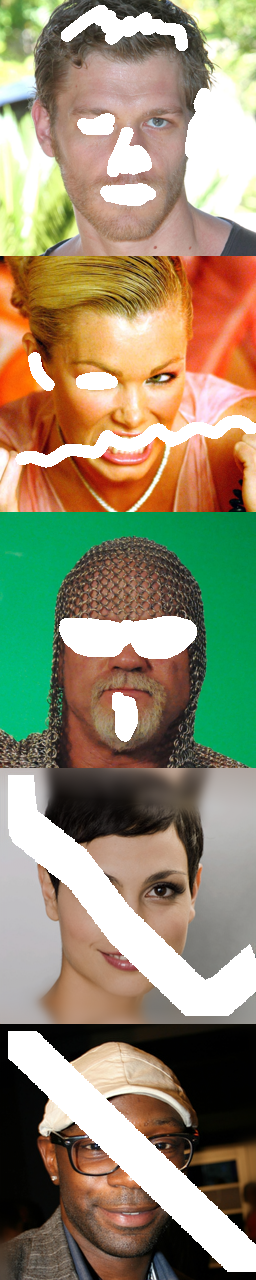}
    \caption{Masked}
  \end{subfigure}%
  \hspace{0.025\textwidth}
  \begin{subfigure}[t]{0.13\textwidth}
      \includegraphics[width=\textwidth]{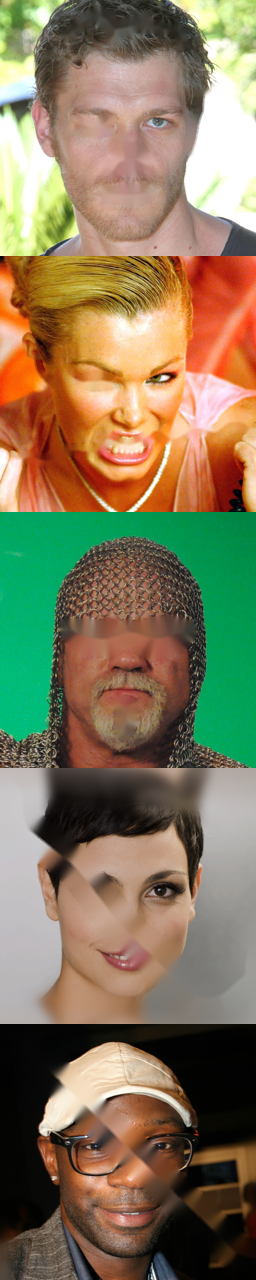}
    \caption{Telea}
  \end{subfigure}%
  \begin{subfigure}[t]{0.13\textwidth}
      \includegraphics[width=\textwidth]{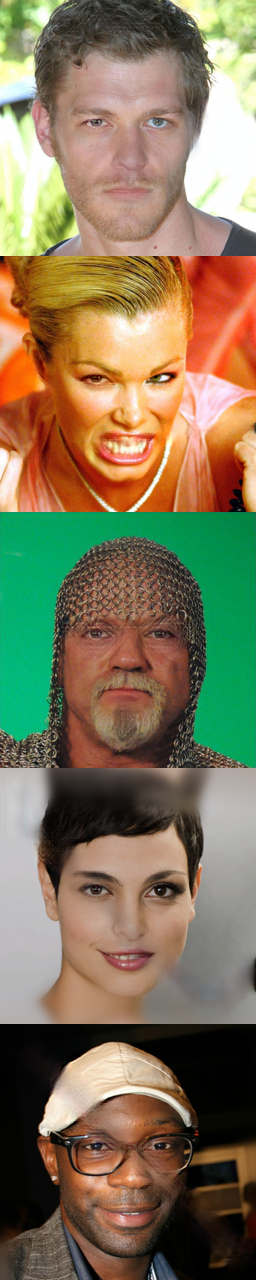}
    \caption{PConv}
  \end{subfigure}% 
  \begin{subfigure}[t]{0.13\textwidth}
      \includegraphics[width=\textwidth]{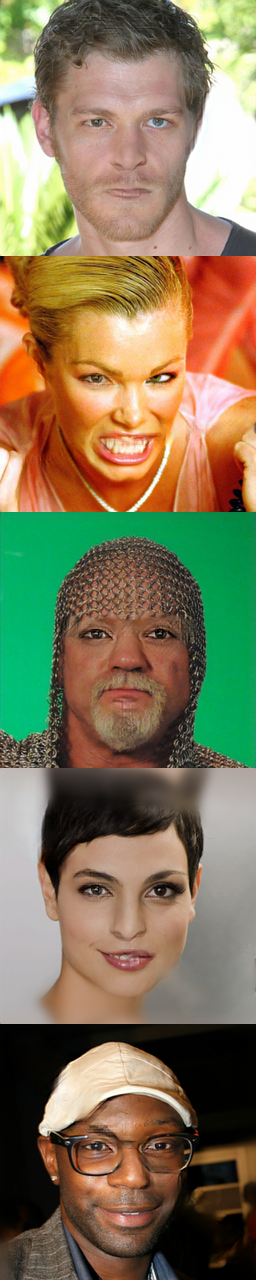}
    \caption{Sim-0}
  \end{subfigure}%
  \begin{subfigure}[t]{0.13\textwidth}
      \includegraphics[width=\textwidth]{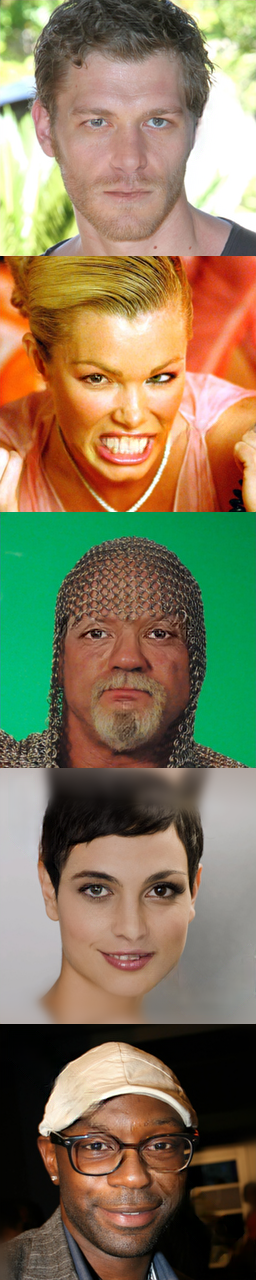}
    \caption{Sim-1 (ours)}
  \end{subfigure}%
  \hspace{0.025\textwidth}
  \begin{subfigure}[t]{0.13\textwidth}
      \includegraphics[width=\textwidth]{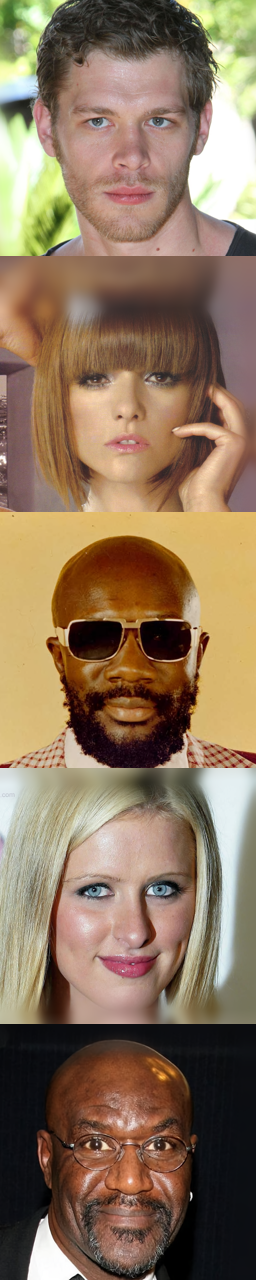}
    \caption{Most similar}
  \end{subfigure}%
\end{center}
   \vspace*{-6mm}
   \caption{Comparison between (c) Telea, (d) PConv, (e, f) our model with 0 and 1 similar image as auxiliary input. (a, b) original and masked input, (g) most similar image which is fed to Sim-1. For details --- zoom in.}
    \label{fig:visual_model_compare}
\end{figure*}

%%%%%%%%% Body TEXT
\section{Introduction}
Inpainting is an important computer vision task, where the goal is to restore masked parts of an image. Inpainting finds wide application in image processing problems like unwanted content removal, red-eye fixing~\cite{yoo2009red} and even eye opening~\cite{dolhansky2017eye}.

Non-learnable methods that use statistics of not masked pieces of the image, like PatchMatch~\cite{barnes2009patchmatch}, produce good inpainting results on images with monotonous or repetitious content (sceneries, textures, \etc). In many ways, this happens due to the fact, that human eye is not sensitive to little discrepancies in images with such content. But inconsistencies in pictures of special domains, like face photos, are very quickly detected by the human eye. This happens because of the so-called "uncanny valley"~\cite{mori1970uncanny} effect.

But recently, deep convolution networks (DCNs) have managed to produce photorealistic results in image inpainting~\cite{iizuka2017globally, liu2018image}. The main difference between DCNs and previous methods is that DCNs learn semantic prior from training data in an end-to-end manner. All the knowledge of DCN is contained in the learned model's parameters and it's convolution architecture, which is a good prior for real-world images, as it has been shown in~\cite{ulyanov2017deep}. But, what if supply DCN with accessory information during inference?

\Eg, for an artist it's easier to draw a portrait of the person, while that person is sitting in front of him, instead of drawing whole painting from memory. An artist can refer to the origin to refine details or to pick correct color. So, we decided to provide DCN with such ability.

In this paper, we focus on image inpainting with auxiliary images. Similar images as auxiliary input can help modern deep neural networks in producing detailed outputs by copying and fitting separate pieces of similar pictures. We suppose that stage of finding similar images is not a barrier, because of rapid progress in large-scale content-based image retrieval~\cite{zhou2017recent} and high availability of a large number of images online.

Another problem which slows down the appearance of inpainting solutions in real-life applications, like smart image editing, is that current research in inpainting is focused on masks with regular forms. Often masks are assumed to be fixed-size squares located in the center of the image~\cite{pathak2016context, yang2017high}. Fortunately, in the recent paper Liu \etal~\cite{liu2018image} proposed method, how to generate irregular masks using two consecutive video frames. However, their dataset has a critical disadvantage --- there is nothing "human" in the utilized source of irregular patterns. Also, this dataset has unnatural sharp crops close to the borders of masks and it's not publicly available.

We propose our irregular mask dataset based on the "Quick, Draw!" dataset~\cite{ha2017neural} (a collection of 50 million human drawings in vector format). We think that the dataset of human drawings is a good source for irregular forms. Combination of strokes from "Quick, Draw!" dataset is effectively infinite, that's why our method can be used as an inexhaustible mask generator. Moreover, the generation process is fully customizable. With bravery, we call it \textbf{"Quick Draw Irregular Mask Dataset"} (QD-IMD).

In summary, our contributions are as follows:
\begin{itemize}
    \item we propose the use of auxiliary images in inpainting task and show that it significantly boosts visual quality
    \item we propose publicly available irregular mask dataset, whose source of irregular patterns is human drawings; introduced mask generation process is fully customizable and effectively inexhaustible
\end{itemize}

\section{Related Work}
Non-learning models often use interpolation methods that rely on pixels of the remaining not-masked parts of an input image. Alexandru Telea developed inpainting technique based on the fast marching~\cite{telea2004image} which is now the default inpainting method in OpenCV~\cite{bradski2000opencv}.
Another popular nonparametric method PatchMatch~\cite{barnes2009patchmatch} iteratively searches for the most suitable patches to fill in the holes. It is one of the state-of-the-art approaches in such class of algorithms.

Deep learning based approaches often use encoder-decoder architecture. Pathak \etal~\cite{pathak2016context} used inpainting task for feature learning with proposed so-called \textit{Content Encoders}.
Yang \etal~\cite{yang2017high}, additionally to \textit{Content Encoders}, uses refinement texture network as a postprocessing stage.
Another popular architecture for inpainting is \textit{Generative Adversarial Network}~\cite{goodfellow2014generative}.
GAN-like architectures managed to solve some problems with blurry and unrealistic outputs~\cite{nguyen2017plug, dolhansky2017eye, isola2017image}.

The recent paper by Liu \etal~\cite{liu2018image} showed stunning results, archiving state-of-the-art in image inpainting for irregular masks. Authors use modification of common convolution layer --- \textit{partial convolution}. Such convolution condition output on valid cells in the mask.

In our research, we use auxiliary images retrieved from the external database. But, an idea of using an external database of images in the inpainting task is not novel. Hays and Efros~\cite{hays2007scene} used a large collection of images to complete missing parts. Another usage of feeding additional data to a network was effectively applied to the closed-to-open eye inpainting in natural pictures~\cite{dolhansky2017eye}. Dolhansky \etal use an extra photo of the same person with opened eyes to consistently inpaint given image. Also, such an approach was successfully applied to image synthesis from semantic maps~\cite{qi2018semi}. Authors combine the parametric and nonparametric techniques, where the nonparametric component is a memory bank of image segments constructed from a training set of images.

\section{Method}
In the classical problem statement of inpainting, masked image and mask are used as input data. Many algorithms use only this information for reconstruction, but in addition, we can retrieve some useful data from the external database to help the algorithm to fill in the missing parts. We're sure that such problem setup is reasonable, because of high availability of a variety of images online and rapid progress in image retrieval techniques.

In our approach, we propose a UNet-like~\cite{ronneberger2015u} neural network similar to~\cite{liu2018image} (but with simple convolution layers) which takes masked image, mask and fixed number of similar images as input ($1$ in our case).

\begin{figure}[h]
\captionsetup[subfigure]{justification=centering}
\begin{center}
    \includegraphics[width=0.48\textwidth]
    {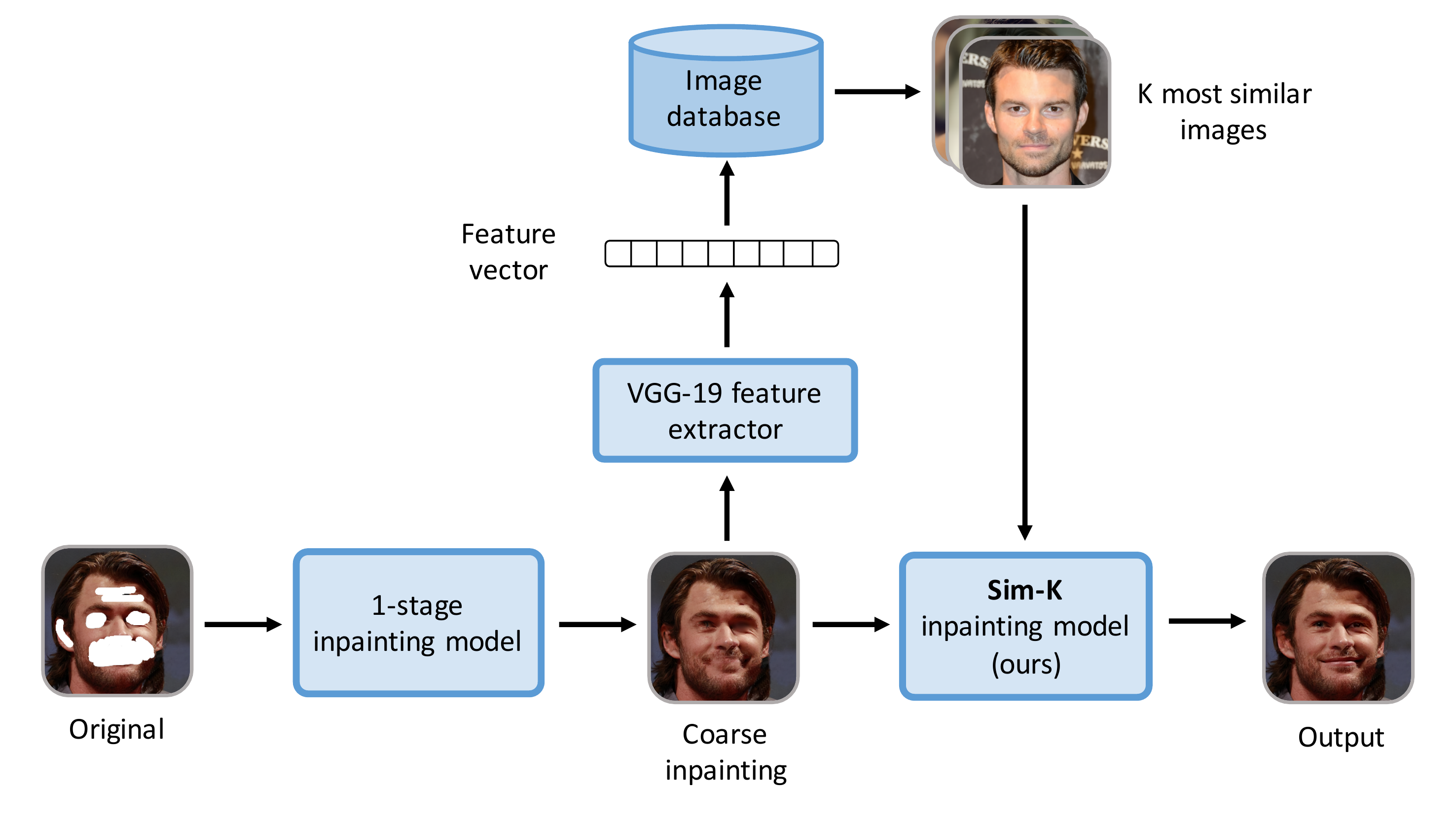}
\end{center}
   \vspace*{-6mm}
   \caption{Schematic representation of our method at test time. In the beginning \textit{Original} masked image is processed by fully-parametric inpainting method to produce \textit{Coarse inpainting} approximation. Then it's fed to the feature extractor and resulting \textit{Feature vector} is used to retrieve \textit{K most similar images} from the image database. In the end, coarse approximation and retrieved similar images are fed to our model Sim-K to get final \textit{Output}.}
    \label{fig:visual_model_compare}
\end{figure}

\subsection{Inpainting with auxiliary examples}
Inpainting task can be formally stated like this: given a masked image $I_{in}$ and corresponding mask $M$, one should reconstruct the initial image $I_{gt}$.

In our approach we additionally find $k$ similar to $I_{in}$ images $\{I^{sim}_{1}, \ldots, I^{sim}_{k}\}$ in the external database, and then use them as extra information.

\subsection{Network design}
Following~\cite{liu2018image} we build similar UNet-like architecture, but instead of using partial convolutions we utilize simple ones. As input we stack masked image $I_{in}$, corresponding mask $M$ and $k$ similar image $\{I^{sim}_{1}, \ldots, I^{sim}_{k}\}$. As a result we obtain a tensor with $3 + 1 + 3 \times k$ channels. 

Unlike~\cite{liu2018image} our input image size is $256 \times 256$ (due to the lack of resources), hence the network's depth is less by one. We don't use batch normalization layers, because they neither accelerate the convergence nor improve final quality in our case.

\subsection{Similar images retrieval}
To retrieve similar images from the database we use kNN-classifier with cosine similarity on images' descriptors. As descriptors we use last convolution layer features from VGG-16~\cite{simonyan2014very} trained on ImageNet~\cite{deng2009imagenet}.

The main difficulty is that input images are corrupted by masks and it's merely impossible to directly extract proper features from them. That's why we use a two-step approach for retrieval: 1) inpaint input masked image with a pre-trained fully-parametric model, 2) find an image in a database similar to the inpainted image. As a pre-trained model, we used an architecture proposed in~\cite{liu2018image}.

During testing, we use described two-step approach, but it's problematically to use it during training due to its long running time. So, for training phase we precalculated top $k$ similar images for all images in the training dataset, using not-masked images for feature extraction. It's not completely fair, but training time reduces dramatically.

\subsection{Loss functions}
We follow~\cite{liu2018image} and use same objectives. Let's say that model's output is $I_{out}$, than per-pixel losses are defined as $L_{hole}={\lVert (1 - M) \odot (I_{out} - I_{gt}) \rVert}_{L_1}$ and $L_{valid}={\lVert M \odot (I_{out} - I_{gt}) \rVert}_{L_1}$.

For calculating perceptual losses we define $I_{comp}$ which is same as $I_{out}$,
but with not-masked pixels explicitly filled with ground truth. We calculate perceptual loss~\cite{gatys2016image} for these two outputs $L^{out}_{perc}$ and $L^{comp}_{perc}$ using features from VGG-16 \textit{pool1}, \textit{pool2} and \textit{pool3} layers.

Than we calculate two style losses~\cite{gatys2016image}: $L^{out}_{style}$ and $L^{comp}_{style}$. We use same layers' outputs from VGG-16 as for perceptual loss.

And the last component is TV-loss~\cite{johnson2016perceptual} $L_{tv}$ calculated for $I_{comp}$ with the $1$-pixel dilation of the hole region.

For the total loss, we sum up all the components with weights. In all experiments we use weights proposed in~\cite{liu2018image}:
\begin{equation}
\begin{split}
    L_{total} = L_{valid} + 6L_{hole} + 0.1L_{tv} + \\
    0.05(L^{out}_{perc} + L^{comp}_{perc}) +\\
    120(L^{out}_{style} + L^{comp}_{style})
\end{split}
\label{eq:ref_total_loss}
\end{equation}

\section{Quick Draw Irregular Mask Dataset}

Many recent approaches focus on rectangular shaped holes, often assumed to be the center of the image. These limitations are absolutely not practical because we often need to inpaint something with irregular form. That's why we need a dataset with masks of irregular forms.

Liu \etal~\cite{liu2018image} proposed such a dataset, where the source of irregular patterns were the results of occlusion/dis-occlusion mask estimation method between two consecutive frames for videos described in~\cite{sundaram2010dense}. That work showed good results in inpainting, but we think their irregular mask dataset has some weaknesses:
\begin{itemize}
  \item There is nothing "human" in generating such masks
  \item Masks often have sharp edges because of rough crops close to borders
  \item It's not public (though authors claimed, they were going to release it)
\end{itemize}

We decided to fight these problems and generated \textbf{Quick Draw Irregular Mask Dataset} (QD-IMD). Our dataset is based on Quick Draw dataset~\cite{ha2017neural} (a collection of 50 million human drawings). Our hypothesis is that a combination of strokes drawn by a human hand is a good source of patterns for irregular masks.

\begin{figure}[h]
\captionsetup[subfigure]{justification=centering}
\begin{center}
  \begin{subfigure}[t]{0.5\textwidth}
      \includegraphics[width=\textwidth]{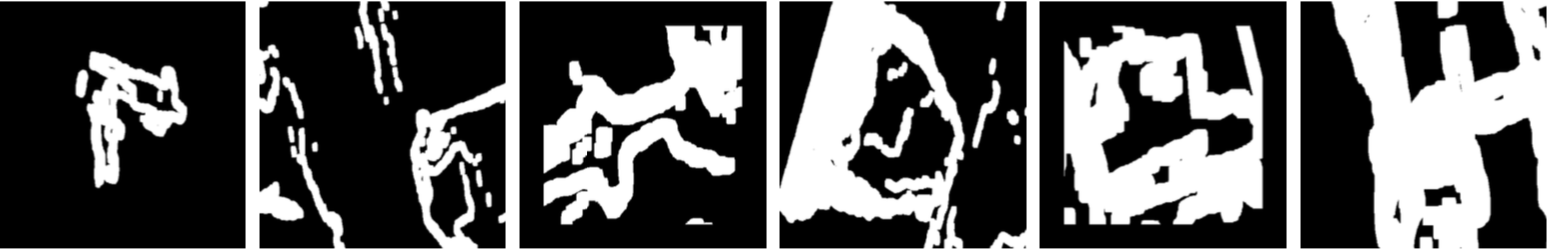}
    \caption{Examples of masks from~\cite{liu2018image} irregular mask dataset}
  \end{subfigure}%
  
  \begin{subfigure}[t]{0.5\textwidth}
      \includegraphics[width=\textwidth]{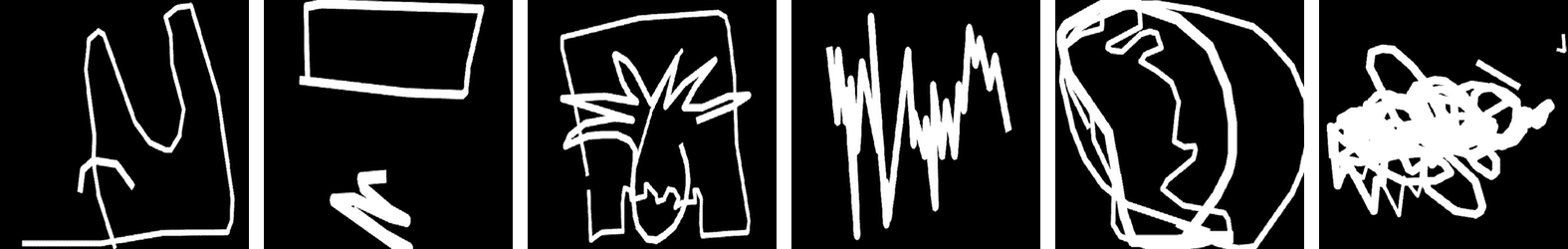}
    \caption{Examples of masks from our QD-IMD}
  \end{subfigure}%
\end{center}
   \vspace*{-6mm}
   \caption{Random masks from (a)~\cite{liu2018image} irregular mask dataset and from (b) our QD-IMD. As it can be seen, masks from our dataset are more natural and smooth.}
\end{figure}

All the parameters of the generation process (number of strokes per image, line width, etc.) are customizable. The number of possible masks is effectively infinite, and the dataset can be used as an inexhaustible mask generator. Algorithm~\ref{alg:qd_imd} shows details of the generation process. Note, that we make a central crop at the end of the procedure to get masks with holes, which touches the boundaries of the image.

\begin{algorithm}[h]
\SetAlgoLined
Randomly choose number of strokes for mask from $\mathcal{N}(4, 2)$\\
Randomly sample strokes from "Quick, Draw" dataset\\
\ForAll{strokes}{
    sample line width (px) from $U(5, 15)$\\
    draw it on the canvas\\
}
Randomly sample upscale rate from $U(1.0, 1.5)$ and upscale canvas correspondingly\\
Make central crop of target shape $256\times 256$\\
Binarize resulting mask\\

 \caption{How QD-IMD is generated}
 \label{alg:qd_imd}
\end{algorithm}

All code and $256 \times 256$ dataset with $50$k training and $10$k testing masks can be found here: \url{https://github.com/karfly/qd-imd}.

\begin{table*}[t]
\begin{center}
\begin{tabular}{l|c|c|c|c|c|c}
                      & $[0.01, 0.1]$ & $(0.1,0.2]$ & $(0.2,0.3]$ & $(0.3,0.4]$ & $(0.4,0.5]$ & $(0.5,0.6]$ \\ \hline
$L_1$(Telea)(\%)&    $28.51\pm6.80$&    $28.47\pm6.87$&    $27.93\pm7.04$&    $28.37\pm6.83$&    $27.07\pm6.17$&    $25.48\pm4.25$    \\
$L_1$(PConv)(\%)&    $0.60\pm0.15$&    $0.93\pm0.30$&    $1.51\pm0.62$&    $3.05\pm0.89$&    $3.99\pm1.07$&    $5.14\pm1.12$    \\
$L_1$(Sim-0)(\%)&    $0.87\pm0.19$&    $1.20\pm0.30$&    $1.75\pm0.51$&    $2.86\pm0.66$&    $3.46\pm0.88$&    $4.92\pm1.14$    \\
$L_1$(Sim-1)(\%)&    $\mathbf{0.55\pm0.17}$&    $\mathbf{0.89\pm0.28}$&    $\mathbf{1.41\pm0.48}$&    $\mathbf{2.52\pm0.66}$&    $\mathbf{3.12\pm0.83}$&    $\mathbf{4.41\pm1.05}$    \\ \hline \hline

PSNR(Telea)&    $9.12\pm2.09$&    $9.15\pm2.28$&    $9.38\pm2.54$&    $9.13\pm1.88$&    $9.49\pm1.81$&    $9.91\pm1.41$    \\
PSNR(PConv)&    $\mathbf{36.94\pm3.16}$&    $\mathbf{32.67\pm2.93}$&    $28.84\pm3.38$&    $23.38\pm2.86$&    $21.54\pm2.55$&    $20.37\pm2.13$    \\
PSNR(Sim-0)&    $35.34\pm2.77$&    $31.52\pm2.58$&    $28.14\pm2.80$&    $24.02\pm2.13$&    $22.76\pm2.21$&    $20.60\pm2.29$    \\
PSNR(Sim-1)&    $36.70\pm3.26$&    $32.41\pm2.79$&    $\mathbf{28.85\pm2.88}$&    $\mathbf{24.58\pm2.40}$&    $\mathbf{23.30\pm2.40}$&    $\mathbf{21.24\pm1.91}$    \\ \hline \hline

SSIM(Telea)&    $0.369\pm0.074$&    $0.371\pm0.079$&    $0.379\pm0.092$&    $0.375\pm0.066$&    $0.365\pm0.065$&    $0.364\pm0.025$    \\
SSIM(PConv)&    $\mathbf{0.986\pm0.006}$&    $\mathbf{0.969\pm0.013}$&    $\mathbf{0.943\pm0.023}$&    $0.884\pm0.030$&    $0.850\pm0.034$&    $\mathbf{0.804\pm0.044}$    \\
SSIM(Sim-0)&    $0.981\pm0.008$&    $0.961\pm0.014$&    $0.932\pm0.024$&    $0.877\pm0.033$&    $0.849\pm0.036$&    $0.783\pm0.043$    \\
SSIM(Sim-1)&    $0.985\pm0.007$&    $0.967\pm0.013$&    $0.939\pm0.023$&    $\mathbf{0.886\pm0.033}$&    $\mathbf{0.858\pm0.037}$&    $0.801\pm0.047$    \\ \hline \hline

IScore(Telea)&    $1.389\pm0.934$&    $1.397\pm0.958$&    $1.399\pm0.976$&    $1.399\pm1.063$&    $1.492\pm1.183$&    $2.276\pm1.185$    \\
IScore(PConv)&    $1.316\pm0.914$&    $1.358\pm0.970$&    $1.402\pm1.069$&    $1.410\pm1.038$&    $1.742\pm1.394$&    $\mathbf{1.387\pm0.800}$    \\
IScore(Sim-0)&    $1.314\pm0.921$&    $\mathbf{1.353\pm0.975}$&    $\mathbf{1.368\pm1.057}$&    $\mathbf{1.367\pm1.029}$&    $1.391\pm0.990$&    $1.600\pm0.578$    \\
IScore(Sim-1)&    $\mathbf{1.308\pm0.908}$&    $1.353\pm0.979$&    $1.370\pm1.058$&    $1.426\pm1.129$&    $\mathbf{1.263\pm0.927}$&    $1.782\pm0.667$    \\ \hline
\end{tabular}
\end{center}
    \caption{Quantitative comparison results on 6 buckets, split according to the hole-to-image area ratio. Our model Sim-1 outperforms other methods comparing with $L_1$ distance and, according to other metrics, our model works better when hole-to-image area ratio is bigger.}
    \label{fig:table_model_compare}
\end{table*}

\section{Experiments}

\subsection{Dataset}
We conduct all experiments on CelebA-HQ~\cite{karras2017progressive} dataset downsampled to $256 \times 256$ size. Faces in this dataset are aligned, so that eyes, mouth and other face parts are located on the same fixed positions. We randomly split it into training ($27$k images) and testing ($3$k images) sets.

Irregular masks for training we generate on the fly using our  QD-IMD mask generator. For testing, we sampled $3$k random masks from QD-IMD test dataset.

We don't use any augmentations. 

\subsection{Training details}
We initialize weights as proposed in~\cite{he2015delving}. For optimization we use Adam~\cite{kingma2014adam} with $0.0001$ learning rate and batch size $12$. All models were trained on two NVidia Tesla K40m GPU (12 Gb) until convergence. Training took about 2 days per one model. As a framework we use PyTorch~\cite{paszke2017automatic}.

\subsection{Comparisons}
We compare 4 methods:
\begin{enumerate}
  \item \textbf{Telea}: non-learning based approach proposed in~\cite{telea2004image}. This method is based on the fast marching technique, which evaluates masked pixels using weighted averages of already estimated pixels in one pass.
  \item \textbf{PConv}: Method proposed by Liu \etal in~\cite{telea2004image}. It's a UNet-like neural network, which uses \textit{partial convolutions} instead of common ones.
  \item \textbf{Sim-0}: Our method with $0$ similar images given.
  \item \textbf{Sim-1}: Our method with $1$ similar image given.
\end{enumerate}

We include \textbf{Telea} in the comparison list, because it is best non-learning based approach available in OpenCV~\cite{bradski2000opencv}. \textbf{PConv} is a state-of-the-art inpainting method, according to results in~\cite{telea2004image}. \textbf{Sim-0} is included into comparison models to clearly demonstrate how visual quality boosts because of using auxiliary information (\textbf{Sim-1}).

\subsection{Quantitative evaluation}
There is still no good metric for numerical evaluation of image inpainting because the task is ill-posed and has many solutions. Following previous papers~\cite{yu2018generative, yang2017high, liu2018image} we calculate 4 metrics to compare methods: 1) $L1$-distance, 2) PSNR, 3) SSIM~\cite{wang2004image} and 4) the inception score (IScore)~\cite{salimans2016improved}.

Table \ref{fig:table_model_compare} shows quantitative results on 6 buckets, split according to the hole-to-image area ratio. You can see that our method \textbf{Sim-1} outperforms other models comparing with $L_1$ distance. According to other metrics, our model works better when hole-to-image area ratio is bigger.

\subsection{User study}
In addition to quantitative metrics, we conducted the side-by-side user study experiment. We randomly chose $30$ images from testing dataset, manually created masks and processed them with $4$ models: Telea, PConv, Sim-1, Sim-0. Then we constructed an online survey with $30$ questions, where a user had to answer which picture seemed most realistic for him. Besides answers, corresponding to all models, there was an available option --- \textbf{"Can't decide"}.

\begin{figure}[t]
\begin{center}
\includegraphics[width=\linewidth]{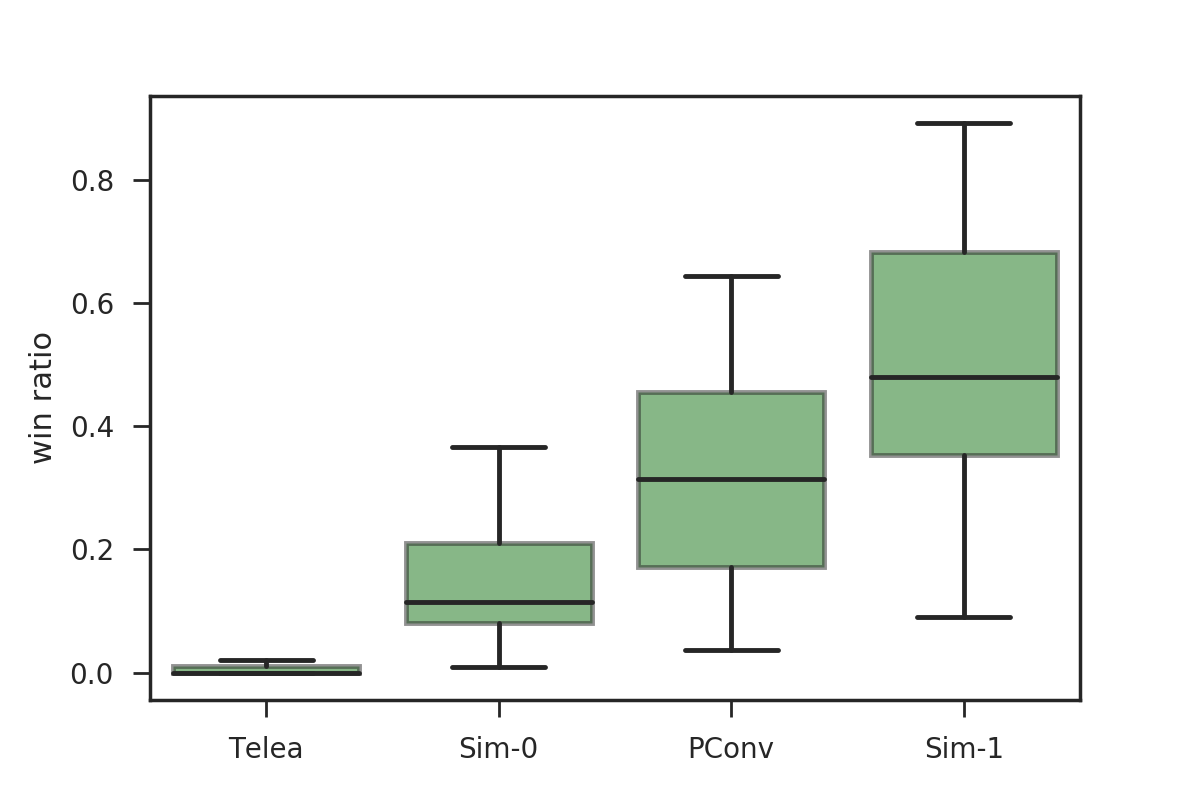}
\end{center}
   \caption{Box plot with win ratio results for each model. Our model Sim-1 is chosen much more frequently, than others.}
\label{fig:model_boxplot}
\end{figure}

$137$ users took part in our experiment. For every question, we calculated the win ratio of each model. Results can be seen in the box plot figure ~\ref{fig:model_boxplot}. On average our model (Sim-1) was chosen by users in $51\%$ of cases. The second most chosen model was PConv with $31\%$ mean win ratio.

In the table~\ref{fig:model_wins} we show the overall number of wins of every model.

\begin{table}[h!]
\begin{center}
\begin{tabular}{|l|c|}
\hline
Method & Wins \\
\hline\hline
Telea & $0 / 30$ \\
PConv & $11 / 30$ \\
Sim-0 & $4 / 30$\\
Sim-1 & $\mathbf{15 / 30}$\\
\hline
\end{tabular}
\end{center}
\caption{Number of wins per each model in user study. Our model Sim-1 wins in most of questions.}
\label{fig:model_wins}
\end{table}

\subsection{Different images as auxiliary input}
To ensure, that Sim-1 model actually uses the auxiliary image as useful extra information for reconstruction, we passed different non-relevant images as auxiliary input.

\begin{figure}[h]
\captionsetup[subfigure]{justification=centering}
\begin{center}
  \begin{subfigure}[t]{0.45\textwidth}
      \includegraphics[width=\textwidth]{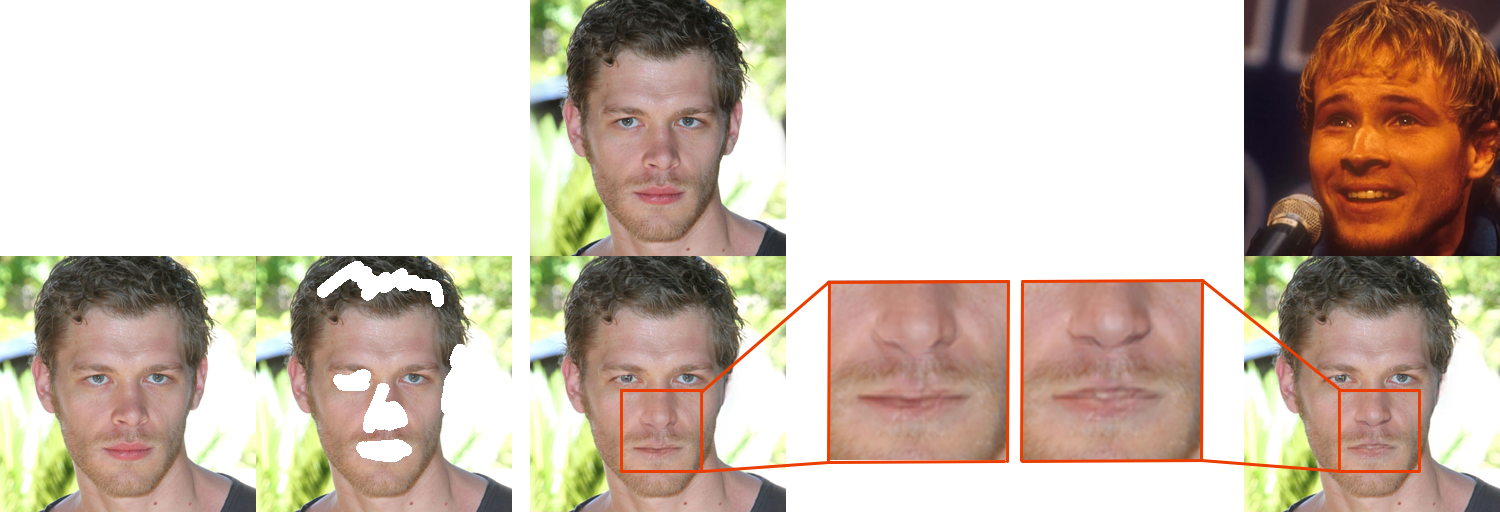}
    \caption{Random image as auxiliary input}
  \end{subfigure}%
  
  \begin{subfigure}[t]{0.45\textwidth}
      \includegraphics[width=\textwidth]{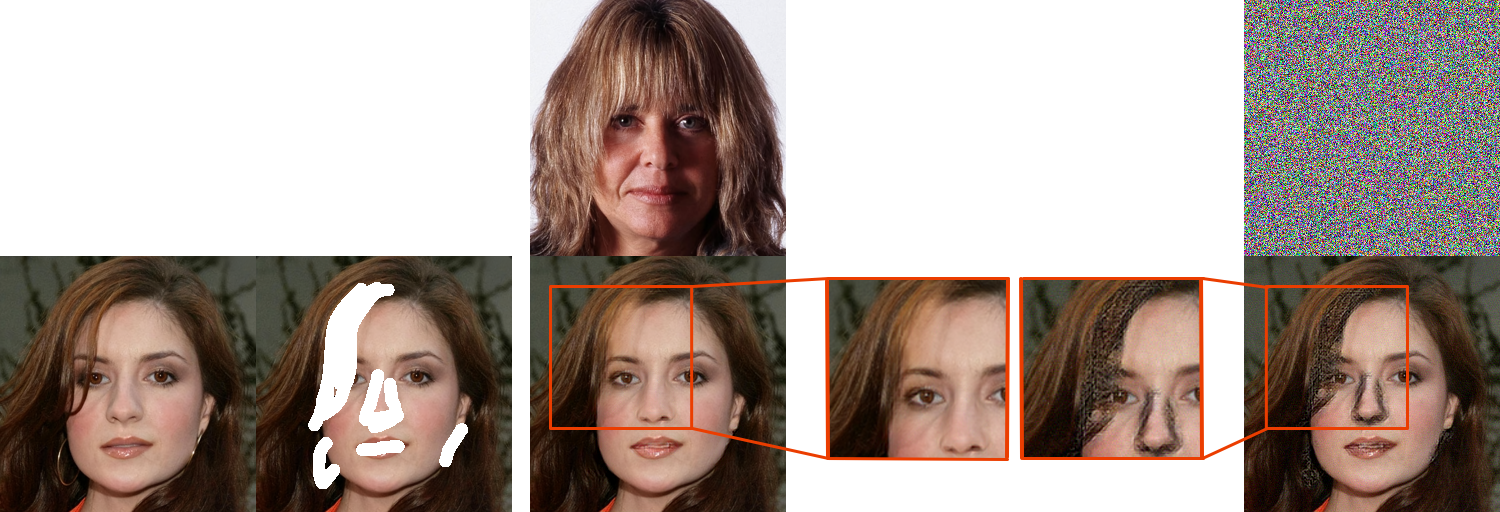}
    \caption{Random noise as auxiliary input}
  \end{subfigure}%
\end{center}
   \vspace*{-6mm}
   \caption{From left to right: original image, masked image, Sim-1 with similar image, Sim-1 with random/noise image. This example shows that our model Sim-1 uses auxiliary input to copy details.}
   \label{fig:compare_nonsimilar}
\end{figure}

First with passed not similar to the input image, but a random picture from the database. In many cases, the output changed a bit, adopting features of the given random picture. You can see an example at figure~\ref{fig:compare_nonsimilar} (a).

Also, we experimented with noise images as auxiliary images. The neural network tries to copy the high-frequency structure of noise to the output, and it results in bad inpainting outcome~\ref{fig:compare_nonsimilar} (b).

These experiments visually show that auxiliary input contributes a lot to the resulting inpainting output.

\subsection{Failure cases}
Our model Sim-1 is not perfect and sometimes it fails to produce reasonable results. In this subsection, we explore some failure cases and try to understand why it happens.

Sim-1 model's failures can be divided into three big categories: blurry results, unrealistic details and inconsistent/asymmetrical outputs. Several examples for each category can be seen in the figure~\ref{fig:sim1_failures}.

Blurriness is a common problem in ill-posed generation tasks. In many cases, blurry results appear because of per-pixel loss terms. GANs successfully deal with this problem, because it's easy for discriminator to distinguish blurry outputs from real.

The $3$-rd and $4$-th rows in the figure~\ref{fig:sim1_failures} show examples of unrealistic results. \Eg in the $4$-th row you can see a creepy eye drawn above the hair. We think this effect appears because of lack of variety in the training dataset. For a neural network it's more difficult to learn context utilization for inpainting, than just spatial information about face parts' locations.

And the last, but not the least issue is related to inconsistent outputs. In the last row in the figure~\ref{fig:sim1_failures}, form and color of the inpainted eye are convenient, but the neural network didn't manage to copy makeup style from the not masked eye. We suppose this problem can be solved by using better image retrieval and using  \textit{global adversarial loss}~\cite{li2018global}.

\section{Conclusions and Future Work}
We found out that additionally feeding similar images to the input of the neural network can be effective in inpainting task. But there are still many dimensions of improvement of such an idea.

In our paper, we chose the simplest network's architecture as a proof-of-concept. We're sure, that better results can be obtained with more appropriate architecture. \Eg it's a good idea to hybridize our approach with~\cite{liu2018image} model that utilizes partial convolutions.

Another side of the algorithm, that can be improved, is similar images retrieval. Though VGG-16 features as images' descriptors produce good results, the quality of descriptors can be significantly boosted by applying, \eg metric learning techniques.

In our research, we didn't tune weights in the total loss. Proper tuning for each model and dataset can considerably improve final visual quality.

Due to the lack of resources, our models were trained only on CelebA-HQ dataset. The future plan is to reproduce the research on other larger datasets like ImageNet.

\clearpage

{\small
\bibliographystyle{ieee}
\bibliography{spii}
}

% \pagebreak

\section*{Appendix}
\begin{figure*}[!hb]
\captionsetup[sub][figure]{justification=centering}
\begin{center}
  \begin{subfigure}[t]{0.15\textwidth}
      \includegraphics[width=\textwidth]{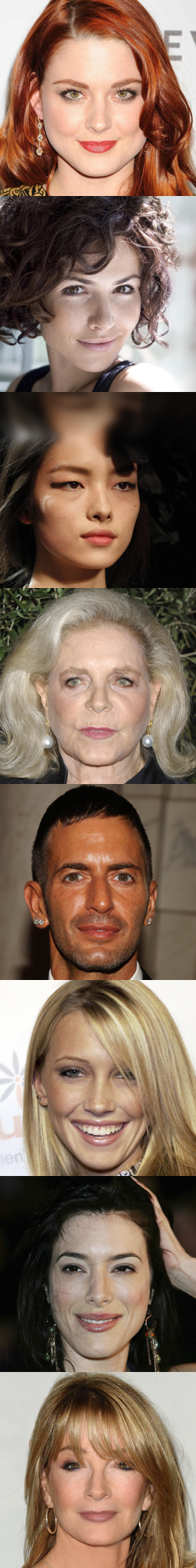}
    \caption{Original}
  \end{subfigure}%
  \begin{subfigure}[t]{0.15\textwidth}
      \includegraphics[width=\textwidth]{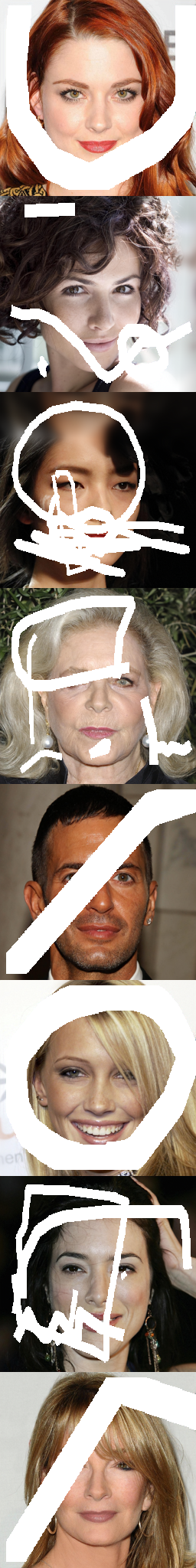}
    \caption{Masked}
  \end{subfigure}%
  \hspace{0.025\textwidth}
  \begin{subfigure}[t]{0.15\textwidth}
      \includegraphics[width=\textwidth]{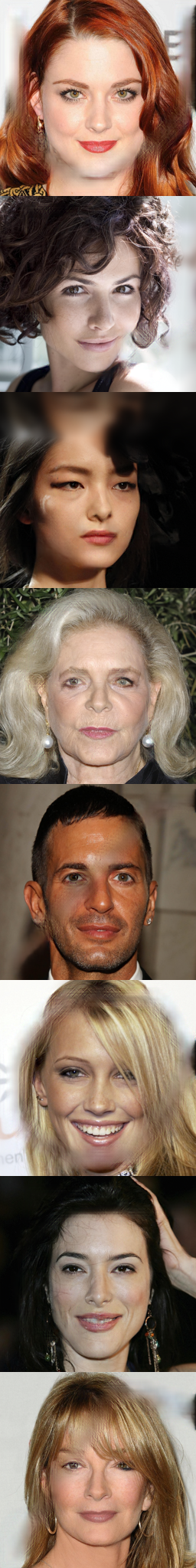}
    \caption{PConv}
  \end{subfigure}% 
  \begin{subfigure}[t]{0.15\textwidth}
      \includegraphics[width=\textwidth]{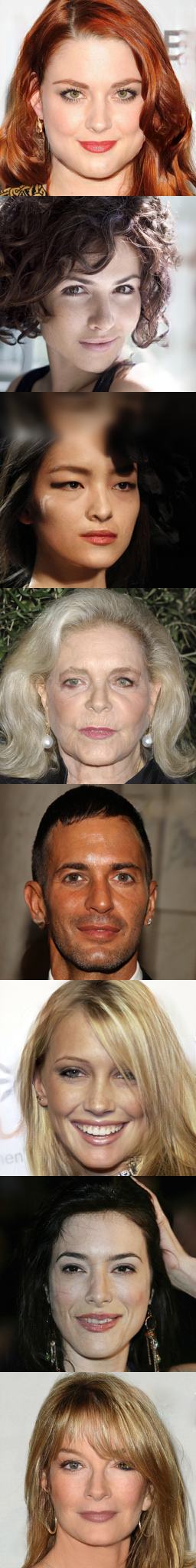}
    \caption{Sim-1 (ours)}
  \end{subfigure}%
  \hspace{0.025\textwidth}
  \begin{subfigure}[t]{0.15\textwidth}
      \includegraphics[width=\textwidth]{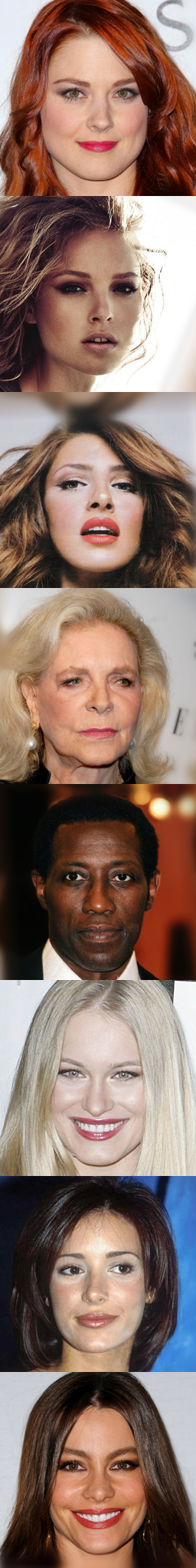}
    \caption{Most similar}
  \end{subfigure}%
\end{center}
   \vspace*{-6mm}
   \caption{Comparison between (c) PConv and (d) our model Sim-1. (a, b) original and masked input, (e) most similar image which is fed to Sim-1. Sim-1 often results in more sharp and realistic details, unlike PConv.}
    \label{fig:compare_pconv_vs_sim1}
\end{figure*}

\begin{figure*}[t]
\captionsetup[subfigure]{justification=centering}
\begin{center}
  \begin{subfigure}[t]{0.19\textwidth}
      \includegraphics[width=\textwidth]{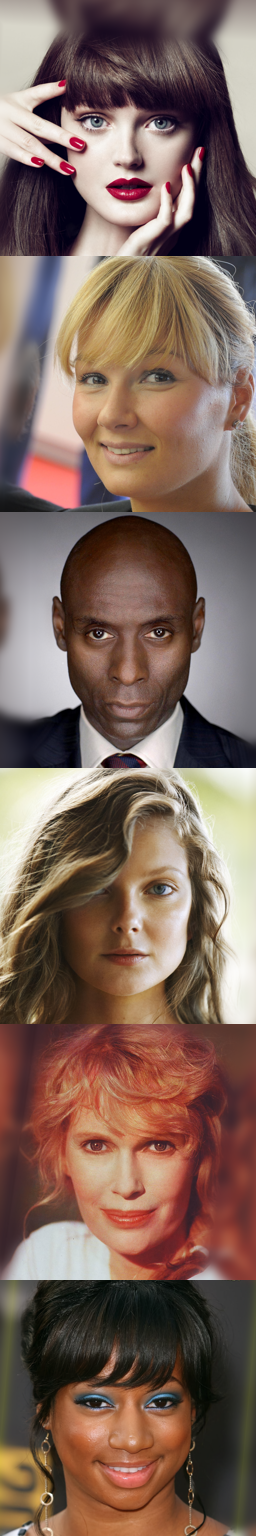}
    \caption{Original}
  \end{subfigure}%
  \begin{subfigure}[t]{0.19\textwidth}
      \includegraphics[width=\textwidth]{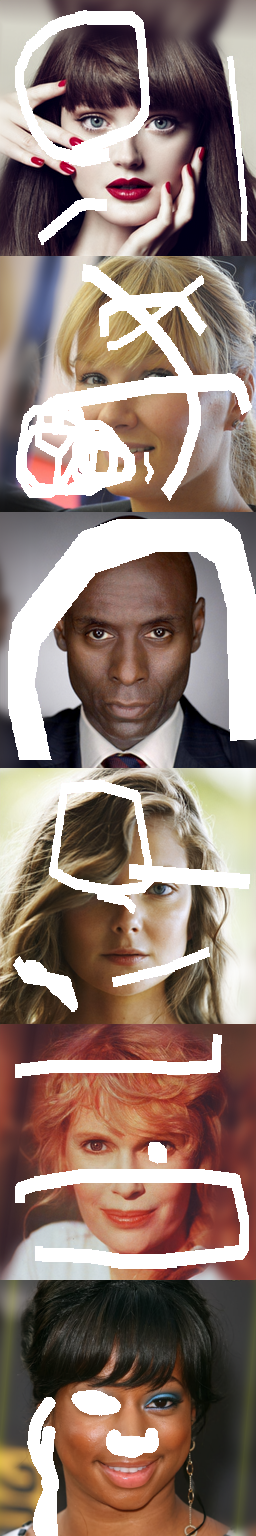}
    \caption{Masked}
  \end{subfigure}%
  \hspace{0.025\textwidth}
  \begin{subfigure}[t]{0.19\textwidth}
      \includegraphics[width=\textwidth]{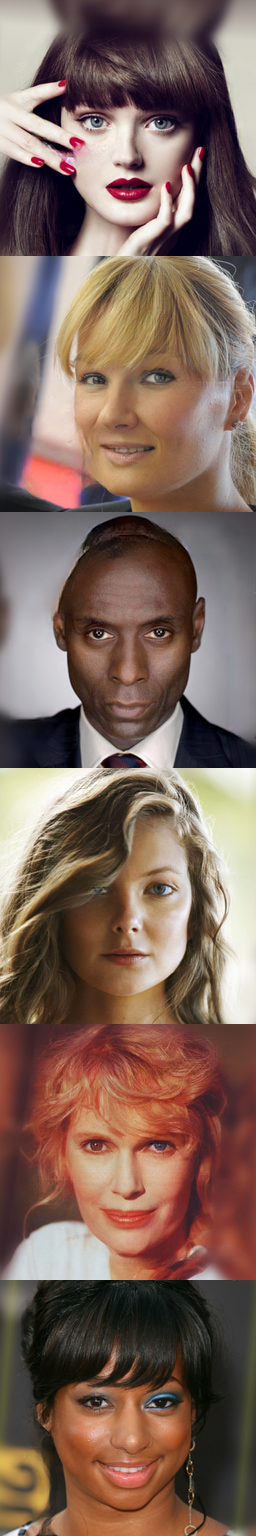}
    \caption{Sim-1 (ours)}
  \end{subfigure}%
  \hspace{0.025\textwidth}
  \begin{subfigure}[t]{0.19\textwidth}
      \includegraphics[width=\textwidth]{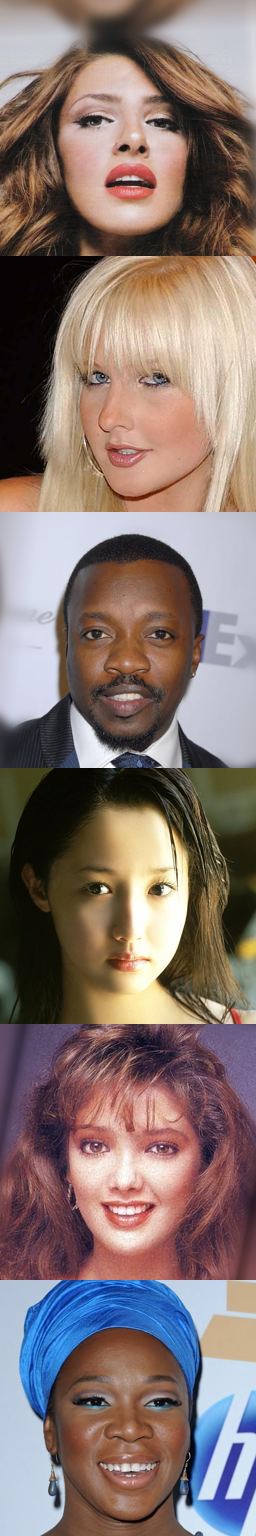}
    \caption{Most similar}
  \end{subfigure}%
\end{center}
   \vspace*{-6mm}
   \caption{Failure cases of (c) our Sim-1 model. (a, b) original and masked input, (d) most similar image which is fed to Sim-1. Sim-1 model's failures can be divided into three groups: (1, 2) blurry results, (3, 4) unrealistic details and (5, 6) inconsistent/asymmetrical outputs.}
    \label{fig:sim1_failures}
\end{figure*}

% \subsection{More examples PConv vs. Sim-1}
\appendix
\renewcommand{\thesubsection}{\Alph{subsection}}
% \part*{Appendix}
In this section we present more visual examples of our model (Sim-1). In the figure~\ref{fig:compare_pconv_vs_sim1} you can see comparison examples between two best models Sim-1 and PConv~\cite{liu2018image}. Additionally in the figure~\ref{fig:sim1_failures} we show failure cases of Sim-1.
% \subsection{More comparison examples PConv vs. Sim-1}

% \subsection{Failure cases}

\end{document}